\documentclass{article}
\usepackage{kr}

\usepackage{times}
\usepackage{soul}
\usepackage{url}
\usepackage[hidelinks]{hyperref}
\usepackage[utf8]{inputenc}
\usepackage[small]{caption}
\usepackage{graphicx}
\usepackage{amsmath}
\usepackage{amssymb}
\usepackage{amsthm}
\usepackage{booktabs}
\usepackage{algorithm}
\usepackage{algorithmic}
\usepackage{xcolor}

\newtheorem{example}{Example}

\newtheorem{definition}{Definition}
\newtheorem{proposition}{Proposition}

\newenvironment{example*}
  {\addtocounter{example}{-1}\example}
  {\endexample}

\newcommand{\PB}[1]{\textcolor{purple}{#1}}

\newcommand{\FT}[1]{\textcolor{
black}{#1}}
\newcommand{\FTT}[1]{\textcolor{black}{#1}}
\newcommand{\del}[1]{\textcolor{olive}{$\times$ #1}}
\newcommand{\AX}[1]{\textcolor{black}{#1}}
\newcommand{\AR}[1]{\textcolor{black}{#1}}
\newcommand{\ARR}[1]{\textcolor{
black}{#1}}
\newcommand{\todo}[1]{\textcolor{magenta}{$*$ #1}}

\newcommand{\trudel}[1]{}
\newcommand{\delete}[1]{\textcolor{olive}{$\times$ #1}}

\def\resp{respectively}
\def\true{\ensuremath{true}}
\def\false{\ensuremath{false}}

\def\SX{SX}
\def\RX{RX}

\def\Exo{\ensuremath{U}}
\def\Endo{\ensuremath{V}}
\def\Eqs{\ensuremath{E}}
\def\fun{\ensuremath{f}}
\def\Var{\ensuremath{W}} 
\def\VarExo{\ensuremath{U}} 
\def\VarEndo{\ensuremath{V}} 
\def\Domain{\ensuremath{\mathcal{D}}} 
\def\val{\ensuremath{w}} 

\def\valendo{\ensuremath{v}} 

\def\Args{\ensuremath{\mathcal{A}}}
\def\Rels{\ensuremath{\mathcal{R}}}
\def\Atts{\ensuremath{\Rels_-}}
\def\Supps{\ensuremath{\Rels_+}}

\def\SF{\ensuremath{\sigma}}
\def\ValSF{\ensuremath{\mathbb{V}}}

\def\sx{\ensuremath{s}}
\def\rx{\ensuremath{r}}
\def\ArgsSX{\ensuremath{\Args^\sx}}
\def\AttsSX{\ensuremath{\Atts^\sx}}
\def\SuppsSX{\ensuremath{\Supps^\sx}}
\def\ArgsRX{\ensuremath{\Args
}}
\def\AttsRX{\ensuremath{\Atts
}}
\def\SuppsRX{\ensuremath{\Supps
}}
\def\ArgsRXacc{\ensuremath{\ArgsRX_a}}

\def\arg{\ensuremath{\alpha}}
\def\arga{\ensuremath{\arg_1}}
\def\argb{\ensuremath{\arg_2}}

\def\relchar{\ensuremath{c}}
\def\Vars{\ensuremath{\mathcal{V}}}
\def\Vara{\ensuremath{\Var_1}}
\def\Varb{\ensuremath{\Var_2}}
\def\Varc{\ensuremath{\Var_3}}

\def\Influences{\ensuremath{\mathcal{I}}}

\def\Input{\ensuremath{\mathbf{u}}}
\def\Inputs{\ensuremath{\mathcal{U}}}

\newcommand{\topdom}[1]{\top_{#1}}
\newcommand{\botdom}[1]{\bot_{#1}}

\title{Explaining Causal Models with Argumentation: \\
the Case of Bi-variate Reinforcement}

\author{%
Antonio Rago$^1$\and
Pietro Baroni$^{2}$\And
Francesca Toni$^1$ \\
\affiliations
$^1$Department of Computing, Imperial College London, UK\\
$^2$Dipartimento di Ingegneria dell'Informazione, Universit\`{a} degli Studi di Brescia, Italy\\
\emails
    \{a.rago, ft\}@imperial.ac.uk, pietro.baroni@unibs.it
}

\begin{document}

\maketitle

\begin{abstract}
Causal models are playing an increasingly important role in machine learning, particularly 
in the realm of explainable AI. We introduce a 
conceptualisation for generating  argumentation frameworks (AFs) from causal models for the purpose of forging explanations for the models' outputs. 
The conceptualisation is based on reinterpreting desirable properties of semantics of AFs as \emph{explanation moulds}, which are means for characterising the relations  
in the causal model argumentatively.
We demonstrate our methodology by 
reinterpreting the property of \emph{bi-variate reinforcement} 
as an explanation mould 
to forge \emph{bipolar AFs} 
as 
explanations for the outputs of 
 causal models. 
We perform a theoretical evaluation of these argumentative explanations, examining whether they satisfy a range of desirable explanatory and argumentative properties. 
\end{abstract}

\section{Introduction}
\label{sec:introduction}

The field of explainable AI (XAI) has in recent years become a major focal point of the efforts of researchers, with a wide variety of models for explanation being proposed (see e.g. \cite{guidotti} for an overview). More recently, incorporating a causal 
perspective into explanations has been explored by some, e.g.
\cite{
Schwab_19,Madumal_20}. 
The link between causes and explanations has long been studied \cite{Halpern_01_UAI}; indeed, the two have even been equated (under a broad sense of the concept of ``cause'') \cite{Woodward_97}. 
Causal reasoning is, in fact, how humans explain to one another \cite{Graaf_17}, and so mimicking such a trend lends credence to the hypothesis that machines should do likewise when their explanations target humans.
Further, research from the social sciences \cite{Miller_19} has indicated the value of causal links, particularly in the form of counterfactual reasoning, within explanations, and that the importance of such information surpasses that of probabilities or statistical relationships for users.
In this paper we outline a methodology for
obtaining explanations from \emph{causal models} \cite{Pearl_99}, based on \emph{(computational) argumentation} (see \cite{AImagazine17,handbook} for recent overviews).


Argumentation 
has received increasing attention in recent years as a means for providing explanations of the outputs of a number of AI models (see \cite{Vassiliades_21,Cyras_21} for recent overviews on argumentative XAI), 
e.g. for recommender systems \cite{Teze_18}, neural classifiers \cite{argflow}, Bayesian networks \cite{Timmer_15} and PageRank \cite{Albini_20}.
\emph{Argumentative explanations} have also been advocated in the social sciences \cite{argExpl,Miller_19}, and several works focus on the power of argumentation to provide a bridge between explained models and users, validated by user studies \cite{Madumal_19,Rago_20}. 
While argumentative explanations 
are wide-ranging in their application, 
their links with causal models 
have remained 
largely unexplored to date.

In this paper, we introduce a conceptualisation for generating \emph{
argumentation frameworks} (AFs) 
from causal models for the purpose of forging explanations for the models' outputs. 
Like \cite{CFX,Albini_21}, we focus  on explaining by  \emph{relations} -- rather than by features, as \AR{is} more conventional (e.g. for feature attribution methods such as~\cite{Lundberg_17})
.
\AR{O}ur method is based on 
a reinterpretation of 
properties of argumentation semantics from the literature 
as \emph{explanation moulds}, i.e. means for characterising argumentative relations (§\ref{sec:forging}).
Here, we 
focus on reinterpreting the property of \emph{bi-variate reinforcement} \cite{Amgoud_18_BAF} 
as a basis 
\AR{for extracting }bipolar AFs \cite{Cayrol:05} which may be used as explanations for the outputs of causal models. 
We  provide a theoretical assessment of these explanations (§\ref{sec:properties}), demonstrating how they satisfy desirable properties from both explanatory and argumentative viewpoints
.

\section{Background}
\label{sec:background}

\AR{Here, w}e provide the core background on causal models and computational argumentation, on which our method relies.

\noindent\textbf{Causal models.}
A \emph{causal model} \cite{Pearl_99} is a triple $\langle \Exo,\Endo,\Eqs\rangle$, where:
$\Exo$ is a (finite) set of \emph{exogenous variables}, i.e. variables whose values are determined by external factors (outside the causal model); 
$\Endo$ is a (finite) set of \emph{endogenous variables}, i.e. variables whose values are determined by internal factors, namely by (the values of some of the) variables 
in $\Exo \cup \Endo$;
each variable may take any value 
in its associated \emph{domain};  we refer to the domain of $\Var_i \in \Exo \cup \Endo$ as $\Domain(\Var_i)$; 
$\Eqs$ is a (finite) set of \emph{structural equations} that, for each  endogenous variable $\VarEndo_i \in \Endo$, define $\VarEndo_i$'s values as a function $\fun_{\VarEndo_i}$ of the values of $\VarEndo_i$'s \emph{parents}
$PA(\VarEndo_i) \subseteq \Exo \cup \Endo \setminus \{\VarEndo_i\}$. 
%
We use the term \emph{binary causal model} to refer to any causal model $\langle \Exo, \Endo, \Eqs \rangle$ such that $\forall \Var_i \in \Exo \cup \Endo$, $\Domain(\Var_i) = \{ 0, 1 \}$ (where $0$ stands for \ARR{``}false\ARR{''} and $1$ stands for \ARR{``}true\ARR{''}, where suitable), and the term \emph{gradual causal model} to refer to any causal model $\langle \Exo, \Endo, \Eqs \rangle$ such that $\forall \Var_i \in \Exo \cup \Endo$, $\Domain(\Var_i)$ is equipped with a partial order (we refer to this partial order as $\leq$
; as usual, $a < b $ stands for $a \leq b$ and $b \nleq a$).
Note that a binary causal model is a gradual causal model (with $0 < 1$).

%
%

Given a causal model $\langle \Exo,\Endo,\Eqs\rangle$ where $\Exo \!= \! \{ \Exo_1, \ldots, \Exo_i \}$,
we denote with $\Inputs$ $=$ $\Domain(\Exo_1) \times \ldots \times \Domain(\Exo_i)$
the set of all possible combinations of values of the exogenous variables (called \emph{
inputs})
.
Abusing
notation, we refer to the value of any 
$\Var_i \in \Exo \cup \Endo$ given 
$\Input \in \Inputs$ as 
$\fun_{\Var_i}[\Input]$
:
if $\Var_i$ is an exogenous variable, $\fun_{\Var_i}[\Input]$ is its assigned value in $\Input$; if $\Var_i$ is an endogenous variable, it is
the value dictated by the structural equations in the causal model.
We use the \emph{do} 
operator \cite{Pearl_12} to indicate 
\emph{interventions}, i.e., for any variable $\VarEndo_i \in \Endo$ 
and value thereof $\valendo_i \in \Domain(\VarEndo_i)$, $do(\VarEndo_{i} = \valendo_i)$ implies that 
$\fun_{\VarEndo_i}$ is replaced by 
the constant function
$\valendo_i$. We use the notation $set(\Var_{i} \!=\! \val_i)$, for $\val_i \!\in\! \Domain(\Var_i)$, to indicate \AR{that} $do(\Var_{i} \!=\! \val_i)$ \AR{if} $\Var_i \in \Endo$ 
 \AR{or} that $\Var_i$ is assigned
$\val_i$ \AR{if} $\Var_i \in \Exo$.  

\noindent \textbf{Argumentation.}
An
\emph{ argumentation framework} (AF)
is any tuple $\langle \Args, \!\Rels_1,\! \ldots, \!\Rels_l \rangle$ with $\Args$ a set (of \emph{arguments}),
and $\Rels_i \!\subseteq \!\Args \!\times\! \Args$, for $i \!\in \!\{1,\ldots,l\}$
, 
 (binary and directed) \emph{dialectical relations} between arguments
\cite{Gabbay:16,prima17}.
In the abstract argumentation
\cite{Dung_95} tradition, arguments in these AFs are unspecified \emph{abstract} entities that can correspond to different concrete instances in different settings. 
Several specific choices of dialectical relations can be made, giving rise to 
specific AFs
, including 
\emph{bipolar AFs} (BFs, our focus in this paper) \cite{Cayrol:05}, 
with $l=2$ and $\Rels_1$ and $\Rels_2$ dialectical relations of \emph{attack} and \emph{support}, \resp, referred to later as \Atts\ and \Supps. 
Given a BF $\langle \Args, \Atts, \Supps \rangle$, for $\arga \in \Args$, we will use the notation $\Atts(\arga) = \{ \argb | (\argb, \arga) \in \Atts \}$ and $\Supps(\arga) = \{ \argb | (\argb, \arga) \in \Supps \}$.
The meaning of BFs (including the intended dialectical role of the relations) may be given in terms of \emph{gradual semantics} 
(e.g. see \cite{prima17,properties}
), defined 
by means of mappings $\SF: \Args \rightarrow \ValSF$, with $\ValSF$ a given set of \emph{values} of interest for evaluating arguments.
The choice of gradual semantics for BFs may be guided by \emph{properties} that $\SF$ should satisfy (e.g. as in \cite{
properties}). 
We will utilise, in §\ref{sec:forging}, a variant of the property of \emph{bi-variate reinforcement} for BFs from \cite{Amgoud_18_BAF}.
We will also use\FT{, in §\ref{sec:properties},}  the following  notions \FT{of coherence} 
from \cite{Cayrol:05}.
Let a \emph{path} from $\arg_x\in \Args$ to $\arg_y\in \Args$ via a relation $\Rels \subseteq \Atts \cup \Supps$  be a sequence of 
arguments $\arg_1, \!\ldots, \!\arg_n$, $n\! \geq \!1$, such that $\arg_1\!=\!\arg_x$, $\arg_n\!=\!\arg_y$, and for each $i$, $1 \!\leq \!i \!< \!n$, $(\arg_n, \! \arg_{n+1}) \!\in \! \Rels$. 
Then, 
a set of arguments $S \!\subseteq \!\Args$ is \emph{internally coherent} iff $\forall \arg_x, \arg_y \!\in \! S$, 
$\nexists$  a path $\arg_1, \!\ldots, \!\arg_n$ from $\arg_x$ to $\arg_y$ via $\Atts \!\cup \!\Supps$  such that $(\arg_{n-1},\arg_n) \!\in \!\Atts$ and for $1 \!\leq \!i \!< n-1$, $(\arg_{i},\arg_{i+1}) \!\in \!\Supps$, 
nor such that $(\arg_{1},\arg_2) \!\in \!\Atts$ and for $2 \!\leq \!i \!< \!n$, $(\arg_{i},\arg_{i+1}) \!\in \!\Supps$.
$S$ is said to be \emph{externally coherent} iff $\forall \arg_x, \arg_y \!\in \!S$, $\nexists \arg_z \!\in \!\Args$ such that
$\exists$ a path from $\arg_x$ to $\arg_z$ via $\Supps$ and
$\exists$ a path $\arg_1, \!\ldots, \!\arg_n$ from $\arg_y$ to $\arg_z$ via $\Atts \! \cup \! \Supps$ such that $(\arg_{n-1},\arg_n) \in \Atts$ and for $1 \!\leq \!i \!< \!n-1$, $(\arg_{i},\arg_{i+1}) \!\in \!\Supps$, or such that $(\arg_{1},\arg_2) \!\in \!\Atts$ and for $2 \!\leq \!i \!< \!n$, $(\arg_{i},\arg_{i+1}) \!\in \!\Supps$.

\section{
From Causal Models to 
Explanation Moulds and Argumentative Explanations}
\label{sec:forging}

We see
the task of obtaining 
\emph{explanations} for causal models' assignments of values to variables 
as a two-step process:
first we
define 
\emph{moulds} characterising the core ingredients of explanations; then we use  these moulds to obtain, automatically, (instances of) AFs as argumentative explanations
.   
Moulds and explanations are defined in terms of \emph{influences} between variables in the causal model, in turn defined in terms of the parent relation underpinning the model
.

\begin{definition}\label{def:influences}
     The  \emph{influence graph} corresponding to a causal model $\langle \Exo,\Endo,\Eqs\rangle$ is the pair $\langle \Vars, \Influences \rangle$, \ARR{where}
        %
        $\Vars = \Exo \cup \Endo$
        \ARR{and}
        $\Influences \subseteq \Vars \times \Vars$ 
        \ARR{such that}
        $\Influences = 
        \{ (\Vara, \Varb) | \Vara \in PA(\Varb)\}$ (referred to as the
        set of \emph{influences}).
\end{definition}

Note that 
influence graph\AR{s} 
\AR{are} closely related to the notion of \emph{causal diagram\AR{s}} \cite{Pearl_95}. While straightforward, 
\AR{they are} useful as 
\AR{they underpin}
much of what follows. 

\begin{figure}[t]
    \centering
    \includegraphics[width=0.45\textwidth]{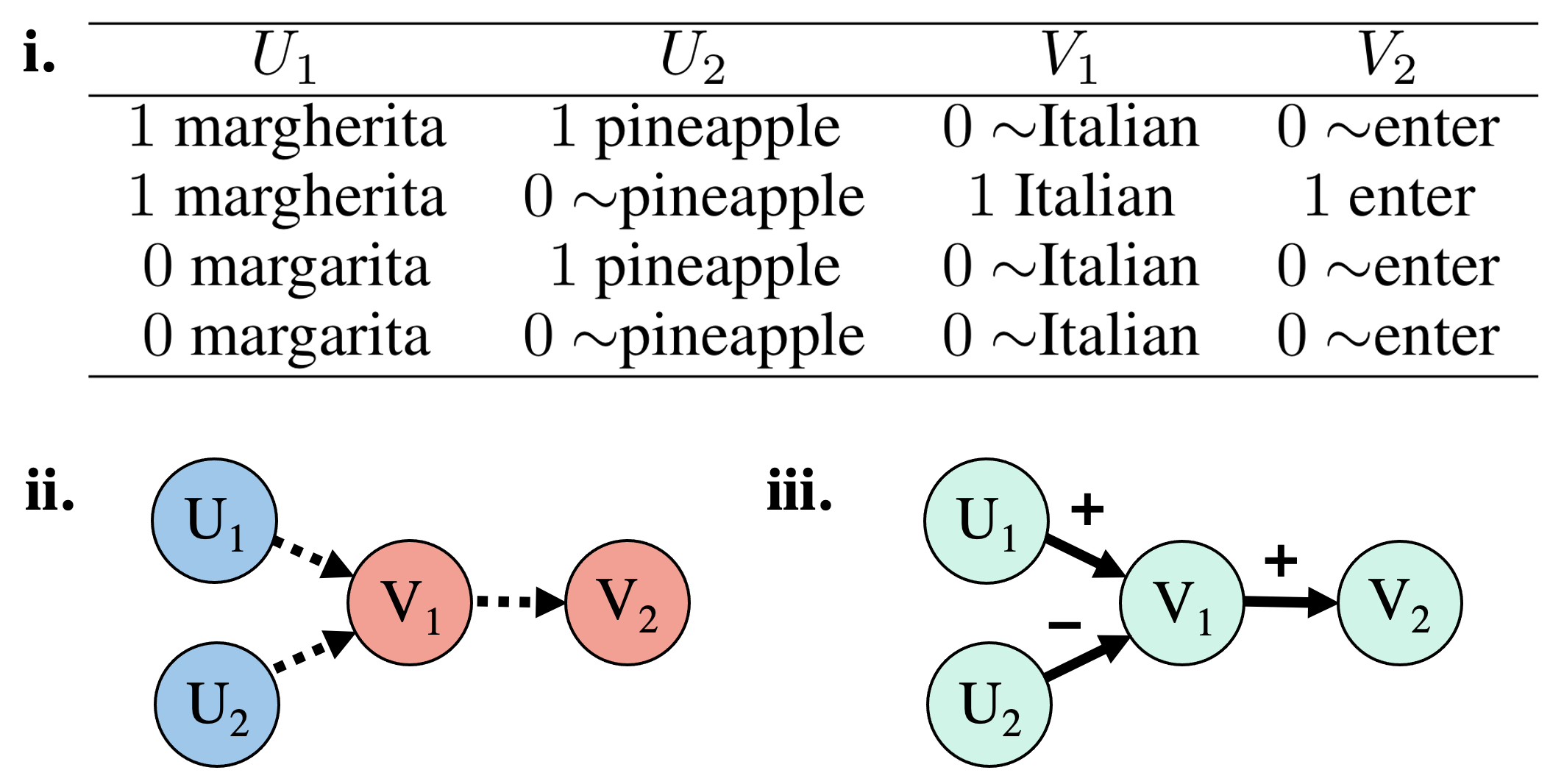}
	\caption{Toy example. (i) Combinations of values ($1$ or $0$) resulting from the structural equations 
	(the assignment of $1$ to $\Exo_1$ may be read as \emph{``margherita'' is spelt correctly on the menu} -- simply given as `margherita' in the table, the assignment of $\Exo_2$ to $0$ may be read as \emph{there is no pineapple on the pizzas} -- simply given as `$\sim $pineapple' in the table, etc.). 
    (ii) Graphical representation of the influence graph,
	with 
	(exogenous/endogenous) variables in the causal model indicated by (blue/red \resp) nodes and influences represented by dashed arrows. (iii) \RX\ 
	for 
	$\Input \in \Inputs$ such that $\fun_{\Exo_1}[\Input] = 1$ and $\fun_{\Exo_2}[\Input] = 0$. 
    \label{fig:pizza}}
\end{figure}

Throughout
, for illustration we will use a toy example 
with a simple (binary) causal model $\langle \Exo,\Endo,\Eqs\rangle$
comprising $\Exo \!\!\!= \!\!\!\{ \VarExo_1, \VarExo_2 \}$, $\Endo \!\!\!=\!\!\! \{ \VarEndo_1, \VarEndo_2 \}$ and ${\forall \Var_i \!\in\! \Exo \!\cup\! \Endo}$, $\Domain(\Var_i) \!\!=\!\! \{ 0, 1 \}$. 
Figure \ref{fig:pizza}i  gives  the combinations of values for the variables
resulting from 
the structural equations $\Eqs$ \FT{(amounting to $\VarEndo_1=\VarExo_1 \wedge \neg \VarExo_2$ and $\VarEndo_2=\VarEndo_1$)} 
and
 Figure \ref{fig:pizza}ii
visualises the influence graph
$\langle \{ \VarExo_1, \! \VarExo_2,\! \VarEndo_1, \!\VarEndo_2 \}, \{ (\VarExo_1,\VarEndo_1), (\VarExo_2,\VarEndo_1), (\VarEndo_1,\VarEndo_2) \}\rangle$
(we ignore Figure \ref{fig:pizza}iii for the moment: this will be discussed later
)
. This 
causal model 
may represent a group's decision on whether 
to enter a restaurant, with variables
$U_1$: \emph{``margherita'' is spelt correctly on the menu, not like the drink};
$U_2$: \emph{there is pineapple on the pizzas};
$V_1$: \emph{the pizzeria seems to be legitimately Italian}; and
$V_2$: \emph{the group chooses to enter the pizzeria}.
Influence graphs synthetically express which variables affect which others but do not give an account of how the influences actually occur in the context the user may be interested in, as expressed by the given values to the exogenous variables. 
For example, the influence graph in Figure~\ref{fig:pizza}ii alone shows which variables affect other variables but provides little intuition on \emph{how} they do so.
Thus, our standpoint is that 
each influence can be assigned an explanatory role, indicating how that influence is actually working in that context.
We assume that each explanatory role is 
specified by a  \emph{relation characterisation}, i.e. a Boolean logical requirement, 
\ARR{that is} 
used 
\ARR{as a}
\emph{mould \ARR{to forge} 
explanations} to be presented to users by indicating which relations play a role 
therein. 

\begin{definition}
    Given a causal model $\langle \Exo,\Endo,\Eqs\rangle$ 
    with corresponding influence graph $\langle \Vars,
    \Influences \rangle$,
	an \emph{explanation mould} is a non-empty set 
	$\{\relchar_1, \!\ldots, \!\relchar_m \}$ where, $\forall i \!\in \!\{1, \ldots, m\}$, $\relchar_i : 
	\Inputs \times \Influences \rightarrow \{ \true, \false \}$ is a \emph{relation characterisation}, in the form of a Boolean condition  
	in some formal language.  
\end{definition}  
Here, we do not prescribe any formal language for specifying 
relation characterisations:  several 
may be suitable. 
\FT{The 
use of this definition requires an up-front choice of the number of relations 
and their characterisations. This choice then applies to all inputs in need of explaining.}

 Given an 
 input \Input, based on an explanation mould 
 we can obtain an AF including, as 
 dialectical relations, the influences satisfying the (different) relation characterisations for the given \Input.
Thus, the choice of relation characterisations is to a large extent dictated by the specific form of 
\emph{argumentative explanation} the intended users expect. 
In general, 
\AR{argumentative explanations} can be generated as follows. 

\begin{definition}\label{def:explanation}
    Given a causal model $\langle \Exo,\Endo,\Eqs\rangle$, its corresponding  influence graph $\langle \Vars, 
    \Influences \rangle$ and an explanation mould 
    $\{ 
    \relchar_1 
    , \ldots, 
    \relchar_m 
    \}$, 
    an \emph{argumentative explanation for $\langle \Exo,\Endo,\Eqs\rangle$ and $\Input \in \Inputs$} is an AF $\langle \Args, \Rels_1, \ldots  \Rels_m \rangle$, where 
    	$ \Args \subseteq 
    	\Vars$, and 
    	$\Rels_1, \ldots,  \Rels_m\subseteq \Influences \cap (\Args \times \Args)$ such that, for any $i=1 \ldots m$, 
    	$\Rels_i=\{(\Vara, \Varb) \in \Influences \cap (\Args \times \Args) | \relchar_i(\Input, (\Vara,\Varb))=\true\}$.
\end{definition}

\FT{Note that these argumentative explanations are \emph{local}, namely they focus on the causal model's behaviour for (any) input $\Input$
. Thus, different argumentative explanations may be obtained for different inputs. 
} 
Note \AR{also} that 
we have left open the choice of $\Args$ (as a generic, possibly non-strict subset of $\Vars$). In practice, $\Args$ may be the full $\Vars$, but we envisage that users may prefer to restrict attention to some variables of interest (for example, excluding variables not
``involved'' in any influence satisfying the relation characterisations).
For example, an argumentative explanation \AR{of a counterfactual nature} for the causal model in Figure \ref{fig:pizza}i and the input in the first row 
may choose to neglect $\VarExo_1$ since changing its value \AR{in this case} 
does not affect the other variables\AR{' values}.

The choice of (number and form of) relation characterisations
in explanation moulds is crucial for the generation of argumentative explanations. 
Here we demonstrate a novel concept for utilising properties of gradual 
semantics for AFs for 
this choice, based on ``property inversion''. %
The idea is to interpret the variable values in the causal model as generated by a ``hypothetical'' gradual semantics embedded in the model itself.
This is similar, in spirit, to recent work to extract (weighted) BFs from multi-layer perceptrons (MLPs) \cite{Potyka_21}, using the underlying computation of the MLPs as a gradual semantics, and to proposals to explain recommender systems 
via tripolar AFs \cite{rec} or BFs \cite{Rago_20},  using the underlying predicted ratings 
as a gradual semantics.
A natural semantic choice for causal models, given that we are trying to explain why endogenous variables are assigned specific values
, given assignments to the exogenous variables,  is to use the assignments as a gradual semantics.

 Then, the  idea of inverting properties of semantics to obtain dialectical relations in AFs  can be recast 
to obtain relation characterisations in explanation moulds as follows:
\emph{given an influence graph and a selected value assignment to exogenous variables, if an influence satisfies 
a given, desirable semantics property, 
then the influence can be cast as part of a dialectical relation with explanatory purposes in the resulting AF}.
We will illustrate this concept with the property of 
\emph{bi-variate reinforcement} for BFs \cite{Amgoud_18_BAF}, which we posit is intuitive in the realm of explanations.
In our formulation of this property, we  require that increasing the value of variables which are attackers (supporters) can only decrease (increase, \resp) the values of variables they attack (support, \resp). 

\begin{definition}\label{def:RX}
Given a gradual causal model 
$\langle \Exo,\Endo,\Eqs\rangle$
and 
influence graph $\langle \Vars, 
\Influences \rangle$,
a \emph{reinforcement explanation mould} is an explanation mould $\{
\relchar_{-}^\rx 
, 
\relchar_{+}^\rx 
\}$ 
such that, given some $\Input \in \Inputs$ 
and 
$(\Vara,\Varb) \in \Influences$: 
\begin{itemize}
    \item $\relchar_{-}^\rx( \Input, (\Vara,\Varb)) = \true$ iff:
    
    \begin{enumerate}
        \item $\forall \val_+ \in \Domain(\Vara)$ such that $\val_+ > 
        \fun_{\Vara}[\Input]$, 
       it holds that $\fun_{\Varb}[\Input, set(\Vara = \val_+)] <_* \fun_{\Varb}[\Input]$ with $<_* = \leq$;
      
        \item $\forall \val_- \in \Domain(\Vara)$ such that $\val_- < 
        \fun_{\Vara}[\Input]$, 
       it holds that $\fun_{\Varb}[\Input, set(\Vara = \val_-)] >_* \fun_{\Varb}[\Input]$ with $>_* = \geq$;
        \item $\exists_{\geq1} \val_+ \in \Domain(\Vara)$ or $\exists_{\geq1} \val_- \in \Domain(\Vara)$ satisfying the conditions at points 1 and 2 with $<_* = <$ and $>_* = >$.
    \end{enumerate}  
    \item $\relchar_{+}^\rx(\Input, (\Vara,\Varb)) = \true$ iff:
    \begin{enumerate}
        \item $\forall \val_+ \in \Domain(\Vara)$ such that $\val_+ > 
        \fun_{\Vara}[\Input]$, 
        it holds that $\fun_{\Varb}[\Input, set(\Vara = \val_+)] >_* \fun_{\Varb}[\Input]$ with $>_* = \geq$;
        \item $\forall \val_- \in \Domain(\Vara)$ such that $\val_- < 
        \fun_{\Vara}[\Input]$, 
        it holds that $\fun_{\Varb}[\Input, set(\Vara = \val_-)] <_* \fun_{\Varb}[\Input]$ with $<_* = \leq$;
        \item $\exists_{\geq1} \val_+ \in \Domain(\Vara)$ or $\exists_{\geq1} \val_- \in \Domain(\Vara)$ satisfying the conditions at points 1 and 2 with $<_* = <$ and $>_* = >$.
    \end{enumerate}  
\end{itemize}
We call any argumentative explanation 
for $\langle \Exo,\!\Endo,\!\Eqs\rangle$ and $\Input
$, given 
a reinforcement explanation mould $\{\relchar_{-}^\rx, \!\relchar_{+}^\rx\}$, a \emph{reinforcement explanation} (\RX)
(for $\langle \Exo,\Endo,\Eqs\rangle$ and $\Input$).
\end{definition}

For illustration, Figure \ref{fig:pizza}iii shows the \RX\ for the causal model in Figure \ref{fig:pizza}i and $\Input$ as in the caption. 
\FT{Note that the causal model can only be understood by inspection of the structural equations; 
instead, the argumentative explanations provide a qualitative characterisation of influences, 
without requiring an understanding of the structural equations.}
Note \ARR{also} that conditions 1 and 2 for the attack and support relations in RXs correspond to a weak form of local monotonicity of the model. For instance, since $\VarExo_2$ 
\AR{attacks}
$\VarEndo_1$, the user knows that, all 
else remaining the same, any increase in the value of $\VarExo_2$ 
cannot give rise to an increase of the value of $\VarEndo_1$, while a decrease of $\VarExo_2$ \AR{will not 
decrease 
the} value of $\VarEndo_1$. 
Condition 3 adds 
a guarantee of effectiveness: there is at least 
one variation of $\VarExo_2$, which, all 
else remaining the same, enforces a variation of $\VarEndo_1$.
Thus RXs have a counterfactual 
\ARR{nature}, as they suggest to the user the kind of local changes with respect to the current situation that could give rise to a desired change of outcome. 
In this respect, 
note that the role assigned to variables refers 
to 
\FT{the selected value assignment
to exogenous variables
}. 
\ARR{For example, in the RX for the input in the first line of the table in Figure \ref{fig:pizza}i, the fact that ``margherita'' is spelt correctly on the menu does not play a role in determining that the pizzeria is not legitimately Italian (indeed this is determined solely by pineapple being on the pizza), thus the support $(\VarExo_1,\VarEndo_1)$ is not present \FTT{in the RX} for this input.}
Such differences reflect the fact that only some (or possibly none) of the individual changes of variables $\VarExo_1$ and $\VarExo_2$ are guaranteed to produce a change \ARR{in $\VarEndo_1$'s value}, depending on the initial context. The local nature of RXs, corresponding to the local nature of bi-variate reinforcement, ensures simplicity and a rather intuitive interpretation but at the same time clearly limits 
expressiveness, in particular RXs are not meant to cover cases where multiple variable changes are needed to 
\ARR{produce} an effect.
Other, 
explanation moulds 
may be needed to satisfy differing users' explanatory requirements.
Note that some 
explanation moulds may be unsuitable to some causal models,  
e.g. our reinforcement explanation mould is not directly applicable to causal models with variables whose domains lack a partial order. 

\section{Properties}
\label{sec:properties}

We perform a theoretical evaluation of RXs with regards to their satisfaction of various 
properties 
(besides \AR{a variant of} bi-variate reinforcement, satisfied by design \AR{and easy to see from Definition \ref{def:RX}}). 
Unless specified otherwise, we assume 
some \RX\ $\langle \ArgsRX, \!\AttsRX,\! \SuppsRX \rangle$ 
for 
gradual causal model $\langle \Exo, \Endo, \Eqs \rangle$ and $\Input \!\in \!\Inputs$ with influence graph $\langle \Vars, \Influences \rangle$.

The first result shows the deterministic nature of RXs.

\begin{proposition}[Uniqueness]\label{prop:uniqueness}
There is no  $\langle \ArgsRX, \AttsRX', \SuppsRX' \rangle$ for 
$\langle \Exo, \Endo, \Eqs \rangle$ and $\Input$, such that $\langle \ArgsRX, \AttsRX', \SuppsRX' \rangle$ is different from $\langle \ArgsRX, \AttsRX, \SuppsRX \rangle$ 
(i.e.  such that 
$\AttsRX' \neq \AttsRX$ or $\SuppsRX' \neq \SuppsRX$).
\end{proposition}

This 
guarantees \emph{stability}~(e.g. as discussed in \cite{Sokol_20}), i.e. a user would never be shown two different explanations for the same causal model 
\FTT{and} input given the choice of $\ArgsRX$. 
We posit that this is important to avoid possible user uncertainty and confusion. 

Further, 
\RX s are acyclic when seen as graphs. 

\begin{proposition}[Acylicity]\label{prop:acyclicity}
Let $(n,e)$ be the graph with $n=\ArgsRX$ and 
$e=\AttsRX \cup \SuppsRX $. 
Then, $(n,e)$ is acyclic.
\end{proposition}

This follows directly from  acyclicity of causal models. It 
prevents \ARR{potentially} undesirable 
behaviour such as a self-attacking or self-supporting 
variable assignments in \RX s
. 

An argument in \RX s may not both attack and support another argument therein:

\begin{proposition}[Unambiguity]\label{prop:clarity}
$\forall \Vara \in \ArgsRX$, $\AttsRX(\Vara) \cap \SuppsRX(\Vara) = \emptyset$, or, equivalently, $\AttsRX \cap \SuppsRX = \emptyset$.
\end{proposition}

Violation of this property would clearly provide contradictory indications to the user.



The following proposition states that argumentative relations in \RX s are derived from causal relationships.

\begin{proposition}[Relevance]\label{prop:relevance}
\AR{$\AttsRX \cup \SuppsRX \subseteq \Influences$. } 
\end{proposition}

Note that, while straightforward for \RX s, this property may be violated by (model-agnostic) explanation methods which do not leverage upon the underlying  causal model.  
This property is in the same spirit as other properties in the XAI literature, e.g. \emph{Dummy} \cite{Sundararajan_20}, which states that a feature which does not affect a classification is given a zero attribution value. 
This may be particularly important in some cases, e.g.  in the running example, it may not be enough to use the absence of pineapple on pizza ($\VarExo_2=0$) as a reason for entering a restaurant ($\VarEndo_2=1$), and 
$\VarEndo_1 = 1$ provides a useful intermediate justification that the restaurant seems to be legitimately Italian.

The following requires that changing attackers or supporters in binary causal models necessitates a change in the value of the argument they attack or support, \resp.

\begin{proposition}[Bipolar Counterfactuality]\label{prop:counterfactuality}
If $\langle \Exo,\Endo,\Eqs\rangle$ is binary, then
$\forall (\Vara,\Varb) \in \Influences$ where $(\Vara,\Varb) \in \AttsRX \cup \SuppsRX$,
for every $\val \neq \fun_{\Vara}[\Input]$: $\fun_{\Varb}[\Input, set(\Vara =\val )] \neq \fun_{\Varb}[\Input]$.
\end{proposition}

This is a powerful explanatory characteristic of \RX s since attacks and supports indicate counterfactuals (in the binary case). 
For example, given the RX in Figure \ref{fig:pizza}iii, a user can immediately see that changing the value of $\VarEndo_2$ can be achieved by changing the value of $\VarEndo_1$, which itself can be achieved by changing the value of $\VarExo_1$ or $\VarExo_2$.

The next 
\FT{property shows}
that behaviour similar to attacks and supports with discrete semantics (see \cite{
Cayrol:05}) 
arises in RXs for binary 
models.

\begin{proposition}[(Dis)agreement]\label{prop:attacks}\label{prop:supports}
If $\langle \Exo,\Endo,\Eqs\rangle$ is binary, then,
$\forall \Vara \in \ArgsRX$: if $\exists \Varb \in \AttsRX(\Vara)$ then 
$\fun_{\Varb}[\Input] \neq \fun_{\Vara}[\Input]$; 
%
if $\exists \Varc \in \SuppsRX(\Vara)$ then
$\fun_{\Varc}[\Input] = \fun_{\Vara}[\Input]$.
\end{proposition}

We thus observe that attacks indicate a contradiction between two arguments while supports indicate harmony between them.
Clearly this is the case in Figure \ref{fig:pizza}, where \ARR{assigning} $\VarExo_2$ 
value $1$ will reduce the value of $\VarEndo_1$ to $0$: a contradiction. Meanwhile, any input which changes $\VarEndo_1$'s value to $0$ will necessitate the same result in $\VarEndo_2$: a harmony.

The set of arguments assigned value $1$ in an \RX\ for a binary causal model 
\FT{satisfies} 
coherence 
\FT{(see §\ref{sec:background})}. 

\begin{proposition}[
(Internal and External) Coherence]\label{prop:acceptability}
If $\langle \Exo,\Endo,\Eqs\rangle$ is binary, then
the set of \emph{accepted arguments} $\ArgsRXacc \!\!=\!\! \{\! \Vara \!\!\!\in\!\! \ArgsRX | \fun_{\Vara}\![\Input] \!\!=\!\! 1 \!\}\!$ is internally and externally coherent.
\end{proposition}

\AR{This result 
indicates} that the basic principles of argumentation are upheld in RXs, which hence can support some genuine forms of argumentative reasoning on the model by the user. 
For example, if the RX 
in Figure \ref{fig:pizza}ii\AR{i} were given for an input $\Input'$ 
\FTT{with} $\fun_{\VarExo_2}[\Input'] = 1$ and $\fun_{\VarEndo_1}[\Input'] = 1$, the set of accepted arguments \FTT{would contain} 
a contradiction\FTT{, } which is not intuitive since the accepted attacker has no effect.

\section{Future Work}\label{sec:conclusions}

%
We believe that 
our approach 
provides the groundwork for 
many future directions. \FT{The computational complexity of RXs deserves attention.} Moulds inspired by other properties, and resulting in other forms of AFs, could be devised
. 
A full empirical analysis of RXs, including 
user studies, 
also seems worthwhile. 
Links between our work and existing XAI methods, particularly those utilising argumentation, could be instructive, while 
counterfactuals and causality also warrant 
investigation in our approach.



\section*{Acknowledgements} Toni was 
 partially funded by the European Research Council (ERC) under the European Union’s Horizon 2020 research and innovation programme (grant agreement No. 101020934). 
Rago and Toni were partially funded by 
J.P. Morgan and by the
Royal Academy of Engineering under the Research Chairs
and Senior Research Fellowships scheme. Any views or opinions expressed herein are solely those of the authors listed
.
The authors would like to thank Fabrizio Russo and Emanuele Albini for their helpful contributions to discussions in the build up to this work.

\bibliographystyle{kr}
\bibliography{bib}

\newpage

\section*{Appendix}

\AR{Here, we give the proofs for all propositions as well as an additional proposition, 
a reformulation of Definition \ref{def:RX}.}

Proof for Proposition \ref{prop:uniqueness}

\begin{proof}
Follows directly from Def. \ref{def:RX}.
\end{proof}

Proof for Proposition \ref{prop:acyclicity}

\begin{proof}
Given that any causal model $\langle \Exo, \Endo, \Eqs \rangle$ (and thus any influence graph $\langle \Vars, \Influences \rangle$ by Def. \ref{def:influences}) is acyclic, any \RX\ must also be acyclic by Def. \ref{def:RX}. 
\end{proof}

Proof for Proposition \ref{prop:clarity}

\begin{proof}
For a proof by contradiction, let $\exists \Vara, \Varb 
\in \ArgsRX$ such that $\Varb \! \in\! \AttsRX(\Vara)\! \cap\! \SuppsRX(\Vara)$.
Then, 
$(\Varb, \!\Vara)$ 
\AX{must satisfy }points 1-3 in Def. \ref{def:RX} for both attacks and supports.
\AX{P}oints 1 and 2 for attacks are compatible with points 1 and 2 for supports 
\AX{iff} $\fun_{\Vara}[\Input, set(\Varb \!=\! \val_-)] \!=\! \fun_{\Vara}[\Input]$ $\forall \val_-$ and $\fun_{\Vara}[\Input, set(\Varb \!=\! \val_+)] \!=\! \fun_{\Vara}[\Input]$ $\forall \val_+$, 
\AX{contradicting} point 3 for 
attacks and supports. Thus, 
\AX{we have a contradiction}.
\end{proof}

Proof for Proposition \ref{prop:relevance}

\begin{proof}
Follows directly from Def. \ref{def:RX}.
\end{proof}

Proof for Proposition \ref{prop:counterfactuality}

\begin{proof}
Since $\Domain(\Vara) = \{0,1\}$, we have four cases: 
1.) $\fun_{\Vara}[\Input_1] = 0$ and $\Vara \in \Atts(\Varb)$;
2.) $\fun_{\Vara}[\Input_1] = 1$ and $\Vara \in \Atts(\Varb)$;
3.) $\fun_{\Vara}[\Input_1] = 0$ and $\Vara \in \Supps(\Varb)$; and
4.) $\fun_{\Vara}[\Input_1] = 1$ and $\Vara \in \Supps(\Varb)$.

For Case 1, since $(\Vara,\Varb) \in \Atts$, Def. \ref{def:RX} gives that either $\exists \val_- \in \Domain(\Vara)$ such that $\val_- < \fun_{\Vara}[\Input]$ or $\exists \val_+ \in \Domain(\Vara)$ such that $\val_+ > \fun_{\Vara}[\Input]$ (since one such value must exist by point 3). The former cannot be true since $\fun_{\Varb}[\Input] = 0$ therefore we can see that $\val_+ = 1$. Then, by Def. \ref{def:RX}, $\fun_{\Varb}[\Input, set(\Vara = \val_+)] < \fun_{\Varb}[\Input]$.
Similarly for Case 2, by Def. \ref{def:RX} we see that $\val_- = 0$ and that $\fun_{\Varb}[\Input, set(\Vara = \val_-)] > \fun_{\Varb}[\Input]$.
For Case 3, by Def. \ref{def:RX} we see that $\val_+ = 1$ and that $\fun_{\Varb}[\Input, set(\Vara = \val_+)] > \fun_{\Varb}[\Input]$.
And finally, for Case 4, by Def. \ref{def:RX} we see that $\val_- = 0$ and that $\fun_{\Varb}[\Input, set(\Vara = \val_-)] < \fun_{\Varb}[\Input]$.
Thus, in all cases $\fun_{\Varb}[\Input, set(\Vara \!\!\neq\!\! \fun_{\Vara}[\Input])] \!\!\neq\!\! \fun_{\Varb}[\Input]$.
\end{proof}

Proof for Proposition \ref{prop:attacks}

\begin{proof}

For a proof by contradiction for disagreement, we let $\fun_{\Vara}[\Input] = \fun_{\Varb}[\Input]$.
We know that $\Domain(\Vara) = \{0,1\}$ since the causal model is binary and so we have two cases.

If $\fun_{\Vara}[\Input]=\fun_{\Varb}[\Input]= 1$ then since $(\Varb,\Vara) \in \Atts$, Def. \ref{def:RX} gives that either $\exists \val_- \in \Domain(\Varb)$ such that $\val_- < \fun_{\Varb}[\Input]$ or $\exists \val_+ \in \Domain(\Varb)$ such that $\val_+ > \fun_{\Varb}[\Input]$ (since one such value must exist). The latter cannot be true since $\fun_{\Varb}[\Input] = 1$ \AX{thus} 
$\val_- = 0$. Then, by Def. \ref{def:RX}, $\fun_{\Vara}[\Input, set(\Varb = \val_-)] > \fun_{\Vara}[\Input]$, which is not possible since we let $\fun_{\Vara}[\Input] = 1$ and so we have a contradiction. 

If $\fun_{\Vara}[\Input]=\fun_{\Varb}[\Input]= 0$ then since $(\Varb,\Vara) \in \Atts$, Def. \ref{def:RX}, by the same process, gives that $\exists \val_+ \in \Domain(\Varb)$ such that 
$\val_+ = 1$. Then, by Def. \ref{def:RX}, $\fun_{\Vara}[\Input, set(\Varb = \val_+)] < \fun_{\Vara}[\Input]$, which is not possible since we let $\fun_{\Vara}[\Input] = 0$ and so we have a contradiction. 

Thus, in both cases 
\AX{we have contradictions.}



For a proof by contradiction for agreement, we let $\fun_{\Vara}[\Input] \neq \fun_{\AX{\Varc}}[\Input]$.
\AX{Again, }we have two cases.

If $\fun_{\Vara}[\Input]= 0$ and $\fun_{\AX{\Varc}}[\Input]= 1$ then since $(\AX{\Varc},\Vara) \in \Atts$, Def. \ref{def:RX} gives that either $\exists \val_- \in \Domain(\AX{\Varc})$ such that $\val_- < \fun_{\AX{\Varc}}[\Input]$ or $\exists \val_+ \in \Domain(\AX{\Varc})$ such that $\val_+ > \fun_{\AX{\Varc}}[\Input]$ (since one such value must exist). The latter cannot be true since $\fun_{\AX{\Varc}}[\Input] = 1$ \AX{thus} 
$\val_- = 0$. Then, by Def. \ref{def:RX}, $\fun_{\Vara}[\Input, set(\AX{\Varc} = \val_-)] < \fun_{\Vara}[\Input]$, which is not possible since we let $\fun_{\Vara}[\Input] = 0$ and so we have a contradiction. 

If $\fun_{\Vara}[\Input]= 1$ and $\fun_{\AX{\Varc}}[\Input]= 0$ then since $(\AX{\Varc},\Vara) \in \Atts$, Def. \ref{def:RX}, by the same process, gives that $\exists \val_+ \in \Domain(\AX{\Varc})$ such that 
$\val_+ = 1$. Then, by Def. \ref{def:RX}, $\fun_{\Vara}[\Input, set(\AX{\Varc} = \val_+)] > \fun_{\Vara}[\Input]$, which is not possible since we let $\fun_{\Vara}[\Input] = 1$ and so we have a contradiction. 

Thus, in both cases 
we have contradictions.
\end{proof}

Proof for Proposition \ref{prop:acceptability}

\begin{proof}
Let us prove the proposition by contradiction.

Regarding internal coherence, let us assume that $\exists \Var_x, \Var_y \in \ArgsRXacc$ such that there exists a path from $\Var_x$ to $\Var_y$ via $\Atts \cap \Supps$ $\Var_1, \ldots, \Var_n$ 
1.) such that $(\Var_{n-1},\Var_n) \in \Atts$ and for $1 \leq i < n-1$, $(\Var_{i},\Var_{i+1}) \in \Supps$, 
or 2.) such that $(\Var_{1},\Var_2) \in \Atts$ and for $2 \leq i < n$, $(\Var_{i},\Var_{i+1}) \in \Supps$. We note that $\forall \Var_z \in \ArgsRXacc$, $\fun_{\Var_z}[\Input] = 1$.

For case 1, $\fun_{\Var_x}[\Input] = 1$ and, by Prop. \ref{prop:supports}, $\fun_{\Var_{n-1}}[\Input] = 1$. Then, Prop. \ref{prop:attacks} requires that $\fun_{\Var_{n}}[\Input] = 0$ which is not possible since $\Var_y \in \ArgsRXacc$ and so we have a contradiction.
For case 2, $\fun_{\Var_x}[\Input] = 1$ and, by Prop. \ref{prop:attacks}, $\fun_{\Var_{2}}[\Input] = 0$. Then, Prop. \ref{prop:supports} requires that $\fun_{\Var_{n}}[\Input] = 0$ which is not possible since $\Var_n \in \ArgsRXacc$ and so we have a contradiction.
Thus, $\ArgsRXacc$ is internally coherent.

Regarding external coherence, $S$ is said to be \emph{externally coherent} iff $\forall \Var_x, \Var_y \in S$,  $\nexists \Var_z \in \Args$ such that 1.)
there exists a path from $\Var_x$ to $\Var_z$ via $\Supps$ and
2.) there exists a path from $\Var_y$ to $\Var_z$ $\Var_1, \ldots, \Var_n$ via $\Atts \cap \Supps$ such that $(\Var_{n-1},\Var_n) \in \Atts$ and for $1 \leq i < n-1$, $(\Var_{i},\Var_{i+1}) \in \Supps$, or such that $(\Var_{1},\Var_2) \in \Atts$ and for $2 \leq i < n$, $(\Var_{i},\Var_{i+1}) \in \Supps$.

Statement 1 requires, by Prop. \ref{prop:supports}, that $\fun_{\Var_z}[\Input] = 1$ since $\fun_{\Var_x}[\Input] = 1$. Meanwhile, both cases for statement 2 require that $\fun_{\Var_z}[\Input] = 0$ by the same logic as the proof for internal coherence, and so we have a contradiction.
Thus, $\ArgsRXacc$ is externally coherent.
\end{proof}

The final proposition, a variant of bi-variate reinforcement
, shows the fundamental nature of attacks and supports \AX{for} gradual causal models: 
strengthening attackers (supporters) can only weaken (strengthen, \resp) an argument.

\begin{proposition}[Bipolar Reinforcement]\label{prop:RX}
$\forall (\Vara,\Varb) \in 
\Influences$, 
$\forall \val_- \in \Domain(\Vara)$ where $\val_- < \fun_{\Vara}[\Input]$, and $\forall \val_+ \in \Domain(\Vara)$ where $\val_+ > \fun_{\Vara}[\Input]$: 
\begin{itemize}
    \item if $(\Vara,\Varb) \in \AttsRX$, then
    $\fun_{\Varb}[\Input, set(\Vara = \val_+)] \leq \fun_{\Varb}[\Input]$ and $\fun_{\Varb}[\Input, set(\Vara = \val_-)] \geq \fun_{\Varb}[\Input]$;
    \item if $(\Vara,\Varb) \in \SuppsRX$, then
    $\fun_{\Varb}[\Input, set(\Vara = \val_+)] \geq \fun_{\Varb}[\Input]$ and $\fun_{\Varb}[\Input, set(\Vara = \val_-)] \leq \fun_{\Varb}[\Input]$.
\end{itemize}
\end{proposition}

\begin{proof}
Follows directly from Def. \ref{def:RX}.
\end{proof}

\end{document}